\documentclass[11pt,a4paper]{article}
\usepackage[hyperref]{acl2018}
\usepackage{times}
\usepackage{latexsym}
\usepackage[justification=centering]{caption}
\usepackage{graphicx}
\usepackage{tabularx}
\usepackage{soul}
\usepackage{textgreek}
\usepackage{multirow}
\usepackage{enumitem}
\usepackage{tabu}
\usepackage{array}
\usepackage{booktabs}
\usepackage{mathtools}
\usepackage{tikz}
\usepackage[normalem]{ulem}
\usepackage{gensymb}
\usepackage{amssymb}
\usepackage{pifont}
\usepackage{tikz}
\usepackage[utf8x]{inputenc}
\usepackage{amsmath}
\usepackage[ruled]{algorithm2e}
\SetAlFnt{\small}
\setlist{noitemsep}
\usepackage{url}

\aclfinalcopy 

\title{Strategies for Language Identification in Code-Mixed Low Resource Languages}

\author{
        Soumil Mandal, Sankalp Sanand \\ \\
       	Department of Computer Science \& Engineering \\ 
        SRM Institute of Science \& Technology, Chennai, India\\
        \textcolor{black!50}{\{soumil.mandal, sankalp.sanand\}@gmail.com}               
}

\date{}

\begin{document}
\maketitle
\begin{abstract}
In recent years, substantial work has been done on language tagging of code-mixed data, but most of them use large amounts of data to build their models. In this article, we present three strategies to build a word level language tagger for code-mixed data using very low resources. Each of them secured an accuracy higher than our baseline model, and the best performing system got an accuracy around 91\%. Combining all, the ensemble system achieved an accuracy of around 92.6\%.
\end{abstract}

\section{Introduction}
On social media, we can often see bilinguals switching back and forth between two languages, a phenomenon commonly referred to as code-switching or code-mixing \cite{sridhar:1980syntax}. For processing such type of textual data, we see that traditional approaches perform quite poorly, especially due to mixing of different rules of grammar, ambiguity, typing variances, as well as other types of noises contributed by informality of the social media aspect. Two types of code-mixing can be commonly seen, one where both the languages share a similar mother script, like English and Spanish. The other being where the mother scripts of both the languages being mixed are different, for example English and Hindi. In the first case, typing is done in the respective native scripts, while in the latter, one of the languages is typed in its transliterated form, so as to maintain homogeneity as well as increase ease of typing. Though there are some standard transliteration rules, for example ISO~\footnote{https://en.wikipedia.org/wiki/ISO\_9}, ITRANS~\footnote{https://en.wikipedia.org/wiki/ITRANS}, it's almost impossible to follow them in real life. Thus, we see variances in transliteration typing, contributed by different phonetic judgments, dialects, informality, etc. which increases the challenges of processing such data even more. Thus, for making any systems for such data, it is extremely important to develop a good language tagger as this will greatly affect the successive modules.

In this article, we present three methods for building language taggers when the amount of code-mixed data available is relatively low. These include usage of convolutional neural networks, data augmentation prior to training, and use of siamese networks. At the end, we have also built an ensemble classifier combining all these 3 methods. The language pairs we worked on are Bengali-English (Bn-En) \& Hindi-English (Hi-En) code-mixed data. Both are Indic languages, having Eastern Nagari Script~\footnote{https://en.wikipedia.org/wiki/Eastern\_Nagari\_script} and Devnagari Script~\footnote{https://en.wikipedia.org/wiki/Devanagari} as their native scripts respectively. Thus, the Bn/Hi tokens are in their phonetically transliterated form in Roman.

\section{Related Work}
In the earlier days, two major strategies were used, n-grams \cite{cavnar:1994n} and dictionary look up \cite{rehurek:2009language}. Several research work has been done in the recent past in order to improve such taggers. \citet{nguyen:2013word} used linear chain CRFs with context information limited to bigrams. \citet{das:2014identifying} utilized multiple features like word level context, dictionary, n-gram, edit distance as features for their classifier. \citet{jhamtani:2014word} created an ensemble model. The first classifier uses edit distance, character n-grams and word frequency, while the second classifier takes output from the first one for current word, along with POS tags of the neighboring tokens. \citet{vyas:2014pos} proposed a method which uses logistic regression and code-switching probability. In the first shared task of code-mixed language identification \cite{solorio:2014overview}, the most popular systems used char n-grams combined with rules, as well as CRFs and HMMs. \citet{piergallini:2016word} used capitalization along with character n-grams. For arbitrary set of languages, a generalized architecture based on HMM was developed by \citet{rijhwani:2017estimating}. \citet{choudhury-EtAl:2017:W17-75} created a model which concatenates character and word embeddings for learning. They also experimented with several curriculum, or order in which the data is presented while training. \citet{mandal:2018language} trained character and phonetic embedding models and then combined them to create an ensemble model. To the best of our knowledge, no work has been done where the amount of data used for building the supervised models is low.

\section{Data Sets}
We collected Bengali (Bn) words from the code-mixed data prepared in \citet{mandal:2018preparing} while Hindi (Hi) words were collected from the data released in ICON 17, tools contest \cite{patra:2018sentiment}. The number of unique tokens collected for both Bn and Hi each was 6000. As the goal is to train our models using low resource, we decided to set the size of our training data at 1000 unique tokens. To have convincing conclusions, as one may argue that the models depend highly on the selected low amount of words, we decided to train multiple models on batches of size 1000 and take the average of the results on our testing data.
\begin{table}[h]
\centering
\scalebox{0.85}{
\begin{tabular}{|c|c|c|c|}
\hline
\textbf{Lang} & \textbf{Train} & \textbf{Dev} & \textbf{Test} \\ \hline
En & 4x1000 & 1000 & 1000 \\ \hline
Bn & 4x1000 & 1000 & 1000 \\ \hline
Hi & 4x1000 & 1000 & 1000 \\ \hline
\end{tabular}}
\caption{Data distribution.}
\label{table 1}
\end{table}

\noindent All the experiments in the following sections were performed individually on 8 sets of data, Bn-En (4) + Hi-En (4), and then average was taken. Also, since testing the final models at instance level is more logical, we extracted random 1000 instances of Bn-En and Hn-En type each for final testing. The mean code-mixing index \cite{das:2014identifying} were 21.4 and 18.7 respectively. 

\section{Baseline System}
As a baseline, we decided to use the character embedding based architecture described in \citet{mandal:2018language}, which uses stacked LSTMs \cite{hochreiter:1997long} of sizes 15-35-15-1, where 15 is the size of input layer, 1 is the size of output layer, and 35, 25 are the sizes of the hidden layers. Batch size was kept at 64 and epochs were set to 100. Optimizer used was Adam \cite{kingma:2014adam}, loss function was binary cross-entropy and activation function for output cell was sigmoid. Bn/Hi was labeled 0 and En as 1. On a whole 8 models were trained (4 for Bn-En and 4 for Hi-En) using distribution shown in Table~\ref{table 1}, tuned using threshold on development data, and tested. Tuning is a simple method used to convert the sigmoid output into a classification, maximizing the accuracy on the development data by altering the threshold using brute force. For example, if \textit{sigmoid\_out} $\geq$ $\theta$, output 1 (En), else 0 (Bn/Hi). The detailed results along with the average is shown in Table~\ref{table 2}.  

\begin{table}[h]
\centering
\scalebox{0.85}{
\begin{tabular}{|c|c|c|c|c|c|}
\hline
\multicolumn{6}{|c|}{\textbf{Bn-En}} \\ \hline
\textbf{Model} & \textbf{$\theta$} & \textbf{Acc} & \textbf{Rec} & \textbf{Prec} & \textbf{F1} \\ \hline
1 &0.87  &80.25  &78.83  &82.70  &80.23  \\ \hline
2 &0.89  &79.05  &77.17  &82.50  &79.02  \\ \hline
3 &0.87  &82.10  &81.10  &83.70  &82.09  \\ \hline
4 &0.86  &81.60  &80.73  &83.00  &81.59  \\ \hline
avg &0.87  &80.75  &79.45  &82.97  &80.73  \\ \hline
\multicolumn{6}{|c|}{\textbf{Hi-En}} \\ \hline
\textbf{Model} & \textbf{$\theta$} & \textbf{Acc} & \textbf{Rec} & \textbf{Prec} & \textbf{F1} \\ \hline
1 &0.84  &80.00  &76.50  &86.60  &79.91  \\ \hline
2 &0.83  &80.40  &76.61  &87.50  &80.30  \\ \hline
3 &0.83  &80.15  &75.62  &86.40  &80.10  \\ \hline
4 &0.84  &80.65  &77.14  &86.45  &80.45  \\ \hline
avg &0.83  &80.30  &76.46  &86.73  &80.19  \\ \hline
\end{tabular}}
\caption{Results of baseline system.}
\label{table 2}
\end{table}

\noindent We can see that the average accuracy achieved for Bn-En is 80.75\% while that for Hi-En is 80.3\%.

\section{Convolutional 1D Network}
Convolutional neural networks \cite{lecun:1999object} have been rapidly gaining popularity in the NLP world, especially as it gives quite satisfactory results, but yet is much faster than recurrent units. In the recent years, CNNs have shown amazing results for text classification problems (\citet{johnson:2014effective}, \citet{johnson:2015semi}). For our experiments, we used the Convolutional 1D variant of CNNs, which is ideal for text. The texts can be thought as images of length 15 (embedding size) and height 1. Thus, n-gram features can be captured using a kernel of size n. 

\subsection{Training}
For implementation we used the Keras \cite{chollet:2015keras} API. The architecture that we employed for training had layers 15-CNN-D(0.2)-CNN-D(0.2)-1. That is, the input layer size is 15, and the output layer size is 1. The hidden layers are CNN units in order, and had filters of size 32. Filters are essentially vectors of weights using which the input is convolved, it provides a measure for how close a patch of input resembles a feature. D(0.2) is the dropout layer, and 0.2 is the dropout rate. This reduces computation cost even further, as well as chances of overfitting. The first CNN unit had kernel size 2 and stride 1 (capturing bigrams), while the second CNN had kernel size 3 and stride 1 (capturing trigrams). Loss function used was binary crossentropy, optimizer used was nadam \cite{sutskever:2013importance}, and activation function for output unit was sigmoid. Batch size was kept at 64 while number of epochs was set to 100 (identical to baseline). Post training, in order to convert the sigmoid outputs to classification, a similar brute force as discussed in baseline system was used.

\subsection{Evaluation}
The results obtained using stacked CNNs are shown below in Table~\ref{table 3}.

\begin{table}[h]
\centering
\scalebox{0.85}{
\begin{tabular}{|c|c|c|c|c|c|}
\hline
\multicolumn{6}{|c|}{\textbf{Bn-En}} \\ \hline
\textbf{Model} & \textbf{$\theta$} & \textbf{Acc} & \textbf{Rec} & \textbf{Prec} & \textbf{F1} \\ \hline
1 &0.89  &84.90  &89.12  &79.50  &84.85  \\ \hline
2 &0.89  &84.85  &89.11  &79.40  &84.80  \\ \hline
3 &0.89  &84.70  &88.90  &79.30  &84.65  \\ \hline
4 &0.88  &84.85  &88.93  &79.60  &84.80  \\ \hline
avg &0.89  &84.83  &89.02  &79.45  &84.78  \\ \hline
\multicolumn{6}{|c|}{\textbf{Hi-En}} \\ \hline
\textbf{Model} & \textbf{$\theta$} & \textbf{Acc} & \textbf{Rec} & \textbf{Prec} & \textbf{F1} \\ \hline
1 &0.88  &84.30  &88.28  &79.10  &84.25  \\ \hline
2 &0.87  &83.75  &87.71  &78.50  &83.71  \\ \hline
3 &0.89  &84.15  &88.15  &78.90  &84.11  \\ \hline
4 &0.88  &83.80  &87.80  &78.50  &83.75  \\ \hline
avg &0.88  &84.00  &87.98  &78.75  &83.96  \\ \hline
\end{tabular}}
\caption{Results using stacked Conv1D.}
\label{table 3}
\end{table}

\noindent A modest improvement can be seen compared to the baseline (about 4\%), which translates to $\approx$ 80 more correct predictions.

\section{Data Augmentation}
Data augmentation has proven to be a useful strategy when data is limited but supervised models are required to be made (\citet{ragni:2014data}, \citet{fadaee:2017data}). The idea we will be employing is to increase the amount of data using probabilistic and neural models which will lead to an increase of feature points. Such augmentation has high chances of extracting new features and essential ones which can boost up the system performance. Two methods were tested for augmenting words in order to increase dataset size, one is a simple n-gram based probabilistic generator, another is a neural generator using LSTM. Both the generator models take two parameters, \textit{seed} and \textit{generation length}. The \textit{seed} is the string which the generator takes as the input in order to generate following characters in order to produce the final word. The \textit{generation length} is the length of the string the generator is asked to generate.

\subsection{N-Gram Generator}
For building the n-gram generator, the classic language modeling approach was taken \cite{ney:1994structuring}, based on Markov assumption. For preparing the data for modeling, the tokens were simply split character wise and stored in an array. Then, the n-gram probabilities were estimated. This is essentially P(\textit{c}\textsubscript{i+1}$\mid$\textit{c}\textsubscript{i}, \textit{c}\textsubscript{i-1}) = \textit{C}(\textit{c}\textsubscript{i+1}, \textit{c}\textsubscript{i}, \textit{c}\textsubscript{i-1})/\textit{C}(\textit{c}\textsubscript{i}, \textit{c}\textsubscript{i-1}), where \textit{C}(.) counts the number of times the given n-gram occurs. We considered character level unigrams, bigrams and trigrams for modeling words, i.e. 3 generators (each for Bn and Hi). For example, the trigram model can be represented as P(\textit{c}\textsubscript{1}, \textit{c}\textsubscript{2}, ..., \textit{c}\textsubscript{n}) $\approx$ P(\textit{c}\textsubscript{1})P(\textit{c}\textsubscript{2}$\mid$\textit{c}\textsubscript{1})$\prod_{i=3}^{n}$P(\textit{c}\textsubscript{i}$\mid$\textit{c}\textsubscript{i-1}, \textit{c}\textsubscript{i-2}). A random function is used to switch randomly between unigram, bigram and trigram. Our n-gram generator had one extra parameter which would control whether the \textit{argmax} character is outputted or the \textit{argsecondmax}, it was determined using a random function. This was done in order to increase variance in generated strings.    

\subsection{LSTM Generator}
In order to increase variance in our augmented data, we decided to make a LSTM based generator. For data preparation, a simple character wise split was done of the existing 1000 tokens. Each word is treated as a separate time series entity. These instances were then numerically mapped with an integer, to create embeddings. Now, simply for preparing the training data, the target is set just by shifting the input sequence by \textit{n} steps. We created 3 generators (each for Bn and Hi), by taking \textit{n} ranging from (1, 3). These generators are essentially learning character based word modeling. Mathematically, given a training sequence (\textit{x}\textsubscript{1}, \textit{x}\textsubscript{2}, ..., \textit{x}\textsubscript{n}), the LSTM uses a sequence of output vectors (\textit{o}\textsubscript{1}, \textit{o}\textsubscript{2}, ..., \textit{o}\textsubscript{n}) to learn a sequence of predictive distributions P(\textit{x}\textsubscript{t+1}$\mid$\textit{x}\textsubscript{$\leq$t}) = \textit{softmax}(\textit{o}\textsubscript{t}). The \textit{softmax} is expressed as P(\textit{softmax}(\textit{o}\textsubscript{t} = \textit{j}) = exp(o\textsubscript{t}\textsuperscript{(j)})/ $\sum_{\textit{k}}$exp(o\textsubscript{t}\textsuperscript{(k)}). Objective of learning is to maximize the total log probability of the training sequence $\sum_{t=0}^{T-1}$log P(\textit{x}\textsubscript{t+1}$\mid$\textit{x}\textsubscript{$\leq$t}). Sampling from this conditional distribution, the next character is generated and is also fed back to the LSTM \cite{sutskever:2011generating}. Our models were implemented using the Keras \cite{chollet:2015keras} API. The layer sizes of our LSTM generator was 26-200-26. Activation function used was softmax, loss was categorical cross-entropy, and optimizer used was rmsprop with a learning rate of 0.001. All other parameters were kept at default. Batch size was kept at 32 and number of epochs was set to 50.  

\subsection{Augmentation}
Before augmentation, we needed to create a list of \textit{seeds} using which the generators would model words and augment data. For this, we simply extracted 1000 substrings, bigram (100), trigram (300), quadgram (600) from training data having highest frequencies. The \textit{generation length} which we decided upon ranged from (3, 16). This value was passed to the generator functions using a random function generating a value between (3, 16). Using these, 1500 tokens was generated by n-gram and LSTM generator each, i.e. a total of 3000 augmented tokens. Combining this with the 1000 training tokens, the number of tokens in the final training data was 4000 for each language.

\subsection{Training}
The architecture and hyperparameters used for training was identical to that of the baseline system. This was done to keep parity, and make the results obtained post augmentation comparable to that of the baseline system. As neural networks are very parameter sensitive, a change in batch size, or any learning functions will force a very different learning making the results incomparable. 

\subsection{Evaluation}
The results of the models trained on augmented data is shown below in Table~\ref{table 4}.
\begin{table}[h]
\centering
\scalebox{0.85}{
\begin{tabular}{|c|c|c|c|c|c|}
\hline
\multicolumn{6}{|c|}{\textbf{Bn-En}} \\ \hline
\textbf{Model} & \textbf{$\theta$} & \textbf{Acc} & \textbf{Rec} & \textbf{Prec} & \textbf{F1} \\ \hline
1 &0.93  &88.45  &87.30  &89.35  &88.44  \\ \hline
2 &0.93  &88.55  &87.60  &89.29  &88.54  \\ \hline
3 &0.94  &88.40  &87.70  &88.94  &88.39  \\ \hline
4 &0.93  &88.50  &87.90  &88.97  &88.49  \\ \hline
avg &0.93  &88.48  &87.63  &89.14  &88.47  \\ \hline
\multicolumn{6}{|c|}{\textbf{Hi-En}} \\ \hline
\textbf{Model} & \textbf{$\theta$} & \textbf{Acc} & \textbf{Rec} & \textbf{Prec} & \textbf{F1} \\ \hline
1 &0.92  &88.20  &87.20  &88.98  &88.20  \\ \hline
2 &0.93  &88.15  &87.00  &89.05  &88.15  \\ \hline
3 &0.93  &88.05  &87.00  &88.87  &88.05  \\ \hline
4 &0.92  &87.90  &86.80  &88.75  &87.89  \\ \hline
avg &0.93  &88.08  &87.00  &88.91  &88.07  \\ \hline
\end{tabular}}
\caption{Results using data augmentation.}
\label{table 4}
\end{table}

\noindent We can see that augmentation has improved the accuracy by a fair amount (about 8\%), which translates to $\approx$ 160 more correctly predicted tokens.

\section{Siamese Network}
Siamese networks \cite{bromley:1994signature} are neural networks whose architecture contains two identical sub-networks, which are joined at a single point, and hence the name. The sub-networks have identical configuration, parameters and weights. In such a network, back-propagation and updating is mirrored across both the networks. A unique property of such networks is that even though it takes two inputs, the order doesn't matter (i.e symmetry). This architecture performs quite well on low resource problems or even one-shot tasks (\citet{koch:2015siamese}, \citet{vinyals:2016matching}). This is because sharing weights across sub-networks results in requirement of fewer parameters to train, which reduces chances of overfitting. They are quite good, and thus popular where finding similarity or relationship between two comparable things is a concern. In the recent past, it has become popular for finding textual similarity as well (\citet{mueller:2016siamese}, \citet{neculoiu:2016learning}, \citet{mou:2016transferable}).
\\ \\
\noindent The siamese network is built around learning a distance metric \textit{dis} between two vectors \textit{x}\textsubscript{1} and \textit{x}\textsubscript{2} (each coming from one sub-network), thus the order doesn't matter, as \textit{dis}(\textit{x}\textsubscript{1}, \textit{x}\textsubscript{2}) should be same as \textit{dis}(\textit{x}\textsubscript{2}, \textit{x}\textsubscript{1}). This key feature is critical to our algorithm. The idea is to consider 1000 Bn/Hi tokens as belonging to one class, and 1000 En tokens belonging to another class. Distance \textit{dis} is assigned following the equation below.
\[   
\textit{dis}(\textit{x}\textsubscript{1}, \textit{x}\textsubscript{2}) = 
     \begin{cases}
       \text{0,} \,\, \textit{x}\textsubscript{1}, \textit{x}\textsubscript{2} \in same class \\
       \text{1,} \,\, \textit{x}\textsubscript{1}, \textit{x}\textsubscript{2} \notin same class \\
     \end{cases}
\]
As a siamese network is trained pair wise, i.e. each vector is paired with every other vector once, with respective targets following the \textit{dis} function, the total training data size increases quadratically. For \textit{Z} members each of \textit{C} classes, which is \textit{Z}$\cdot$\textit{C} vectors on a whole, the total number of possible pairs is given by
\[
    \textit{N}\textsubscript{pairs} = \binom{\textit{Z}\cdot\textit{C}}{2} = \frac{(\textit{Z}\cdot\textit{C})!}{2!(\textit{Z}\cdot\textit{C}-2)!}
\]
In our case, \textit{Z} = 1000, \textit{C} = 2, which results in a total of 19,99,000 vector pairs for our training data. Thus, we can see a huge increment in the size of training data, thus reducing the chances of overfitting, which is a big problem for neural networks, especially when amount of data is low like in our case.

\subsection{Architecture}
For each of the sub-networks of our siamese network, the layer sizes were 15-128-128-64-D(0.1)-32-D(0.1)-16. Here, 15 is the size of the input layer, while 16 is the size of the output layer. 128-128-64-32 are the sizes of the hidden layers in order. The layers of sizes 128 were GRU cells \cite{cho:2014learning}, while 64-32 were dense networks with feed forward cells. D(0.1) is a dropout layer with rate 0.1 added in between the layers 64-32 and 32-16, in order to reduce overfitting further more, which additionally reduces computing cost as well. The two tensors of sizes (1, 16) from each of the subnetworks are then concatenated, and sent to the output cell. This model was implemented using Tensorflow \cite{abadi:2016tensorflow}.

\subsection{Training}
The whole model was trained to minimize contrastive loss which is given by the equation below
\[
    (1-\textit{Y})\frac{1}{2}(D\textsubscript{w})^2 + (\textit{Y})\frac{1}{2}\{\textit{max(0, m $-$ (D\textsubscript{w})}\}^2
\]
Where \textit{D}\textsubscript{w} is defined as the euclidean distance between the outputs from the two sub-networks, which mathematically can be represented as the equation below. $\lambda$ denotes a very small value, which in our case was 10\textsuperscript{-6}.
\[
    \textit{D}\textsubscript{w} = \sqrt[]{\{\textit{G}\textsubscript{w}(\textit{X}\textsubscript{1}) - \textit{G}\textsubscript{w}(\textit{X}\textsubscript{2})\}^2 + \lambda}
\]
An L2 weight decay \cite{van:2017l2} term was also used in the loss to encourage the network to learn less noisy weights and improve generalization. The activation function used for the output cell was sigmoid, and optimizer used was rmsprop with a learning rate of 0.0001 (other values kept at default). Batch size was kept at 128, and number of epochs was set to 10. 

\subsection{Tuning}
Since our siamese model gives a sigmoid out as well, thus, to convert it into a classifier we had to find a threshold for which the accuracy is maximum. This was done using the development data. A random selection of 100 tokens was done from training data (Bn/Hi) to create our \textit{support set}. Support set is the data with which the network compares the inputs to give a similarity score. Our algorithm essentially takes an input token, compares it with tokens in our support set, and stores the similarity scores in an array. The sum of this array is considered as the final similarity score. As our classifiers output is sigmoid, the range of the sum of this array is (0, 100). This process was done for 2x1000 tokens of our development data. Finally, using a similar brute force technique used for the other models, the $\theta$ for which the accuracy was maximum on the development was calculated (shown in Table~\ref{table 5}).

\subsection{Evaluation}
For evaluation on our testing data, the algorithm that was used is described below.
\begin{algorithm}[h]
 \KwData{token to be tagged (\textit{token})}
 \KwResult{language tag}
 similarities = [], similarity = 0 \\
 \For{support in support\_set}{
  similarity = siamese(\textit{support}, \textit{token})\;
  similarities.append(similarity)\;
 }
 \eIf{sum(similarities) $\leq$ $\theta$}{
   return 0\;
   }{
   return 1\;
  }
 \caption{Siamese language tagger}
\end{algorithm}

\noindent Using the above algorithm, the results on our testing data is shown in Table~\ref{table 5}.

\begin{table}[h]
\centering
\scalebox{0.85}{
\begin{tabular}{|c|c|c|c|c|c|}
\hline
\multicolumn{6}{|c|}{\textbf{Bn-En}} \\ \hline
\textbf{Model} & \textbf{$\theta$} & \textbf{Acc} & \textbf{Rec} & \textbf{Prec} & \textbf{F1} \\ \hline
1 &22.7  &89.45  &87.80  &90.79  &89.45  \\ \hline
2 &22.8  &89.55  &87.90  &90.89  &89.55  \\ \hline
3 &22.8  &89.85  &88.10  &91.29  &89.85  \\ \hline
4 &22.7  &89.45  &88.10  &90.54  &89.45  \\ \hline
avg &22.8  &89.58  &87.98  &90.88  &89.56  \\ \hline
\multicolumn{6}{|c|}{\textbf{Hi-En}} \\ \hline
\textbf{Model} & \textbf{$\theta$} & \textbf{Acc} & \textbf{Rec} & \textbf{Prec} & \textbf{F1} \\ \hline
1 &22.6  &89.50  &88.00  &90.72  &89.49  \\ \hline
2 &22.6  &89.35  &87.80  &90.61  &89.35  \\ \hline
3 &22.7  &89.65  &88.30  &90.75  &89.65  \\ \hline
4 &22.6  &89.50  &88.10  &90.64  &89.49  \\ \hline
avg &22.6  &89.50  &88.05  &90.68  &89.50  \\ \hline
\end{tabular}}
\caption{Results using siamese network.}
\label{table 5}
\end{table}
\noindent We can see that the siamese architecture has performed quite well as expected, and have achieved accuracies above 89\% for both Bn-En and Hi-En. This is an improvement of 9\% from our baseline system, i.e. it has correctly predicted $\approx$ 180 tokens more. Also, unlike the other systems, here the accuracy difference between the Bn-En and Hi-En has reduced quite a bit (0.03\%). This is probably due to the fact that the threshold was calculated over 100 instances, which normalizes variances to quite an extent.

\section{Ensemble Model}
As we had created multiple systems for language tagging, we decided to ensemble them using simple weighted voting technique. It's basically bootstrap aggregating \cite{hothorn:2003double} with different weights, based on their accuracies. Weight \textit{w}\textsubscript{n} associated with classifier \textit{c}\textsubscript{n} can be calculated as \textit{a}\textsubscript{n}/$\sum_{i=0}^{n}a\textsubscript{i}$, where a\textsubscript{n} is the respective accuracies on testing data. Thus, $\sum_{i=0}^{n}w\textsubscript{i} = 1$. The class having the greater weight is given as the final output. Using this method, and  averaging over all the four data sets, the results obtained are shown in Table~\ref{table 6}.
\begin{table}[h]
\centering
\scalebox{0.85}{
\begin{tabular}{|c|c|c|c|c|}
\hline
\textbf{Lang} & \textbf{Acc} & \textbf{Rec} & \textbf{Prec} & \textbf{F1} \\ \hline
Bn-En &89.95  &88.90  &90.80  &89.95  \\ \hline
Hi-En &89.80  &88.50  &90.86  &89.79  \\ \hline
\end{tabular}}
\caption{Averaged ensemble results.}
\label{table 6}
\end{table}

\noindent A slight improvement in accuracy can bee seen, but is still quite important as the tokens are all unique. The accuracy scores achieved, 89.95\% and 89.80\% are very close to that of \citet{mandal:2018language}, which reports an accuracy of 91.71\% using character embedding, where $\approx$ 6.6k tokens were utilized for training (6.6x times data used for building our models). Also, recall and precision both have close values, suggesting the model is not biased towards a single language. This is unlike the individual systems, and is quite a good aspect.   	
\section{Instance Level Testing}
To get an estimate on how our trained models will perform in real applications, we tested our trained and tuned models at instance level. This is also important as it maybe the case that the model has failed to learn very commonly used tokens for a certain language, for example articles like 'the', 'an', etc which will result in a much reduced accuracy when tested at instance level. The best performing classifiers from each of the methods were chosen for testing. The results obtained are shown below in Table~\ref{table 7}.

\begin{table}[h]
\centering
\scalebox{0.85}{
\begin{tabular}{|c|c|c|c|c|}
\hline
\multicolumn{5}{|c|}{\textbf{Bn-En}} \\ \hline
\multicolumn{1}{|c|}{\textbf{Method}} & \multicolumn{1}{l|}{\textbf{Acc}} & \multicolumn{1}{c|}{\textbf{Rec}} & \multicolumn{1}{l|}{\textbf{Prec}} & \multicolumn{1}{c|}{\textbf{F1}} \\ \hline
Conv1D &88.15  &85.80  &90.03  &88.14  \\ \hline
Augment &90.85  &89.60  &91.89  &90.85  \\ \hline
Siamese &91.00  &90.99  &92.53  &90.99  \\ \hline
Ensemble &92.65  &91.90  &93.29  &92.65  \\ \hline
\multicolumn{5}{|c|}{\textbf{Hi-En}} \\ \hline
\textbf{Method} & \multicolumn{1}{l|}{\textbf{Acc}} & \multicolumn{1}{c|}{\textbf{Rec}} & \multicolumn{1}{l|}{\textbf{Prec}} & \multicolumn{1}{c|}{\textbf{F1}} \\ \hline
Conv1D &88.10  &85.90  &89.85  &88.09 \\ \hline
Augment &90.25  &88.40  &91.79  &90.25  \\ \hline
Siamese &91.05  &89.20  &92.62  &91.05  \\ \hline
Ensemble &92.60  &91.20  &93.83  &92.59  \\ \hline
\end{tabular}}
\caption{Results on instance level testing.}
\label{table 7}
\end{table}

\noindent From the table, we can see that the all the three models have performed quite well at instance level as well, proving their effectiveness. Here, the impact of the ensemble model is more visible as well, as it as achieved the highest accuracy by $>$ 1\% margin.

\section{Summary \& Discussion}
All the three methods that have been demonstrated, have their own sets of advantages and disadvantages. Though the siamese network has got the overall best scores, the classifier is relatively slow, having a tag time of $\approx$ 1
sec per~\footnote{i5-6300HQ (2.8 GHz) + 16GB RAM} token. The Conv1D has the fastest tagging time, but is relatively inferior in terms of performance. Augmentation method performs almost as good as the siamese method, plus is faster, but creating two generators for augmentation is a slightly tedious work. One interesting thing that can be noticed from all the experiments is that for all 4x2 datasets, the results are quite close, suggesting that above a certain amount, the weights learned by the network are quite similar. Recall and precision values are also pretty close to each other, which tells us the architectures doesn't create very biased models. Also, other than Conv1D, in the other methods (including baseline), we can see precision $>$ recall. This suggests that the models generalized the Indic character patterns quite well, which results in higher number of true positives.

\section{Conclusion \& Future Work}
In this article, we present experimental results from our ongoing work on building word level language taggers for code-mixed data when the amount of resources is quite low, which is actually the case many a times. We present three effective methods for this, which are Conv1D networks, data augmentation and siamese. Over all, siamese networks gave the best results with an accuracy of 91\%. Conv1D and data augmentation performed quite satisfactorily as well and gave accuracies of 88\% and 90\% respectively. We also create an ensemble classifier combining these, which achieves an accuracy of 92.6\%. In future, we would like to explore more methods, especially using memory augmented neural networks \cite{santoro:2016one}. Using transfer learning (\citet{pratt:1993discriminability}, \citet{do:2006transfer}) may prove to be an effective solution as well.
\bibliography{acl2018}

\begin{thebibliography}{38}
\expandafter\ifx\csname natexlab\endcsname\relax\def\natexlab#1{#1}\fi

\bibitem[{Abadi et~al.(2016)Abadi, Barham, Chen, Chen, Davis, Dean, Devin,
  Ghemawat, Irving, Isard et~al.}]{abadi:2016tensorflow}
Mart{\'\i}n Abadi, Paul Barham, Jianmin Chen, Zhifeng Chen, Andy Davis, Jeffrey
  Dean, Matthieu Devin, Sanjay Ghemawat, Geoffrey Irving, Michael Isard, et~al.
  2016.
\newblock Tensorflow: a system for large-scale machine learning.
\newblock In \emph{OSDI}, volume~16, pages 265--283.

\bibitem[{Bromley et~al.(1994)Bromley, Guyon, LeCun, S{\"a}ckinger, and
  Shah}]{bromley:1994signature}
Jane Bromley, Isabelle Guyon, Yann LeCun, Eduard S{\"a}ckinger, and Roopak
  Shah. 1994.
\newblock Signature verification using a" siamese" time delay neural network.
\newblock In \emph{Advances in neural information processing systems}, pages
  737--744.

\bibitem[{Cavnar et~al.(1994)Cavnar, Trenkle et~al.}]{cavnar:1994n}
William~B Cavnar, John~M Trenkle, et~al. 1994.
\newblock N-gram-based text categorization.
\newblock \emph{Ann arbor mi}, 48113(2):161--175.

\bibitem[{Cho et~al.(2014)Cho, Van~Merri{\"e}nboer, Gulcehre, Bahdanau,
  Bougares, Schwenk, and Bengio}]{cho:2014learning}
Kyunghyun Cho, Bart Van~Merri{\"e}nboer, Caglar Gulcehre, Dzmitry Bahdanau,
  Fethi Bougares, Holger Schwenk, and Yoshua Bengio. 2014.
\newblock Learning phrase representations using rnn encoder-decoder for
  statistical machine translation.
\newblock \emph{arXiv preprint arXiv:1406.1078}.

\bibitem[{Chollet et~al.(2015)}]{chollet:2015keras}
Fran\c{c}ois Chollet et~al. 2015.
\newblock Keras.
\newblock \url{https://keras.io}.

\bibitem[{Choudhury et~al.(2017)Choudhury, Bali, Sitaram, and
  Baheti}]{choudhury-EtAl:2017:W17-75}
Monojit Choudhury, Kalika Bali, Sunayana Sitaram, and Ashutosh Baheti. 2017.
\newblock \href {http://www.aclweb.org/anthology/W/W17/W17-7509} {Curriculum
  design for code-switching: Experiments with language identification and
  language modeling with deep neural networks}.
\newblock In \emph{Proceedings of the 14th International Conference on Natural
  Language Processing (ICON-2017)}, pages 65--74, Kolkata, India. NLP
  Association of India.

\bibitem[{Das and Gamb{\"a}ck(2014)}]{das:2014identifying}
Amitava Das and Bj{\"o}rn Gamb{\"a}ck. 2014.
\newblock Identifying languages at the word level in code-mixed indian social
  media text.

\bibitem[{Do and Ng(2006)}]{do:2006transfer}
Chuong~B Do and Andrew~Y Ng. 2006.
\newblock Transfer learning for text classification.
\newblock In \emph{Advances in Neural Information Processing Systems}, pages
  299--306.

\bibitem[{Fadaee et~al.(2017)Fadaee, Bisazza, and Monz}]{fadaee:2017data}
Marzieh Fadaee, Arianna Bisazza, and Christof Monz. 2017.
\newblock Data augmentation for low-resource neural machine translation.
\newblock \emph{arXiv preprint arXiv:1705.00440}.

\bibitem[{Hochreiter and Schmidhuber(1997)}]{hochreiter:1997long}
Sepp Hochreiter and J{\"u}rgen Schmidhuber. 1997.
\newblock Long short-term memory.
\newblock \emph{Neural computation}, 9(8):1735--1780.

\bibitem[{Hothorn and Lausen(2003)}]{hothorn:2003double}
Torsten Hothorn and Berthold Lausen. 2003.
\newblock Double-bagging: Combining classifiers by bootstrap aggregation.
\newblock \emph{Pattern Recognition}, 36(6):1303--1309.

\bibitem[{Jhamtani et~al.(2014)Jhamtani, Bhogi, and
  Raychoudhury}]{jhamtani:2014word}
Harsh Jhamtani, Suleep~Kumar Bhogi, and Vaskar Raychoudhury. 2014.
\newblock Word-level language identification in bi-lingual code-switched texts.
\newblock In \emph{Proceedings of the 28th Pacific Asia Conference on Language,
  Information and Computing}.

\bibitem[{Johnson and Zhang(2014)}]{johnson:2014effective}
Rie Johnson and Tong Zhang. 2014.
\newblock Effective use of word order for text categorization with
  convolutional neural networks.
\newblock \emph{arXiv preprint arXiv:1412.1058}.

\bibitem[{Johnson and Zhang(2015)}]{johnson:2015semi}
Rie Johnson and Tong Zhang. 2015.
\newblock Semi-supervised convolutional neural networks for text categorization
  via region embedding.
\newblock In \emph{Advances in neural information processing systems}, pages
  919--927.

\bibitem[{Kingma and Ba(2014)}]{kingma:2014adam}
Diederik~P Kingma and Jimmy Ba. 2014.
\newblock Adam: A method for stochastic optimization.
\newblock \emph{arXiv preprint arXiv:1412.6980}.

\bibitem[{Koch et~al.(2015)Koch, Zemel, and Salakhutdinov}]{koch:2015siamese}
Gregory Koch, Richard Zemel, and Ruslan Salakhutdinov. 2015.
\newblock Siamese neural networks for one-shot image recognition.
\newblock In \emph{ICML Deep Learning Workshop}, volume~2.

\bibitem[{van Laarhoven(2017)}]{van:2017l2}
Twan van Laarhoven. 2017.
\newblock L2 regularization versus batch and weight normalization.
\newblock \emph{arXiv preprint arXiv:1706.05350}.

\bibitem[{LeCun et~al.(1999)LeCun, Haffner, Bottou, and
  Bengio}]{lecun:1999object}
Yann LeCun, Patrick Haffner, L{\'e}on Bottou, and Yoshua Bengio. 1999.
\newblock Object recognition with gradient-based learning.
\newblock In \emph{Shape, contour and grouping in computer vision}, pages
  319--345. Springer.

\bibitem[{Mandal et~al.(2018{\natexlab{a}})Mandal, Das, and
  Das}]{mandal:2018language}
Soumil Mandal, Sourya~Dipta Das, and Dipankar Das. 2018{\natexlab{a}}.
\newblock Language identification of bengali-english code-mixed data using
  character \& phonetic based lstm models.
\newblock \emph{arXiv preprint arXiv:1803.03859}.

\bibitem[{Mandal et~al.(2018{\natexlab{b}})Mandal, Mahata, and
  Das}]{mandal:2018preparing}
Soumil Mandal, Sainik~Kumar Mahata, and Dipankar Das. 2018{\natexlab{b}}.
\newblock Preparing bengali-english code-mixed corpus for sentiment analysis of
  indian languages.
\newblock \emph{arXiv preprint arXiv:1803.04000}.

\bibitem[{Mou et~al.(2016)Mou, Meng, Yan, Li, Xu, Zhang, and
  Jin}]{mou:2016transferable}
Lili Mou, Zhao Meng, Rui Yan, Ge~Li, Yan Xu, Lu~Zhang, and Zhi Jin. 2016.
\newblock How transferable are neural networks in nlp applications?
\newblock \emph{arXiv preprint arXiv:1603.06111}.

\bibitem[{Mueller and Thyagarajan(2016)}]{mueller:2016siamese}
Jonas Mueller and Aditya Thyagarajan. 2016.
\newblock Siamese recurrent architectures for learning sentence similarity.
\newblock In \emph{AAAI}, volume~16, pages 2786--2792.

\bibitem[{Neculoiu et~al.(2016)Neculoiu, Versteegh, and
  Rotaru}]{neculoiu:2016learning}
Paul Neculoiu, Maarten Versteegh, and Mihai Rotaru. 2016.
\newblock Learning text similarity with siamese recurrent networks.
\newblock In \emph{Proceedings of the 1st Workshop on Representation Learning
  for NLP}, pages 148--157.

\bibitem[{Ney et~al.(1994)Ney, Essen, and Kneser}]{ney:1994structuring}
Hermann Ney, Ute Essen, and Reinhard Kneser. 1994.
\newblock On structuring probabilistic dependences in stochastic language
  modelling.
\newblock \emph{Computer Speech \& Language}, 8(1):1--38.

\bibitem[{Nguyen and Do{\u{g}}ru{\"o}z(2013)}]{nguyen:2013word}
Dong Nguyen and A~Seza Do{\u{g}}ru{\"o}z. 2013.
\newblock Word level language identification in online multilingual
  communication.
\newblock In \emph{Proceedings of the 2013 Conference on Empirical Methods in
  Natural Language Processing}, pages 857--862.

\bibitem[{Patra et~al.(2018)Patra, Das, and Das}]{patra:2018sentiment}
Braja~Gopal Patra, Dipankar Das, and Amitava Das. 2018.
\newblock Sentiment analysis of code-mixed indian languages: An overview of
  sail\_code-mixed shared task@ icon-2017.
\newblock \emph{arXiv preprint arXiv:1803.06745}.

\bibitem[{Piergallini et~al.(2016)Piergallini, Shirvani, Gautam, and
  Chouikha}]{piergallini:2016word}
Mario Piergallini, Rouzbeh Shirvani, Gauri~S Gautam, and Mohamed Chouikha.
  2016.
\newblock Word-level language identification and predicting codeswitching
  points in swahili-english language data.
\newblock In \emph{Proceedings of the Second Workshop on Computational
  Approaches to Code Switching}, pages 21--29.

\bibitem[{Pratt(1993)}]{pratt:1993discriminability}
Lorien~Y Pratt. 1993.
\newblock Discriminability-based transfer between neural networks.
\newblock In \emph{Advances in neural information processing systems}, pages
  204--211.

\bibitem[{Ragni et~al.(2014)Ragni, Knill, Rath, and Gales}]{ragni:2014data}
Anton Ragni, Kate~M Knill, Shakti~P Rath, and Mark~JF Gales. 2014.
\newblock Data augmentation for low resource languages.
\newblock In \emph{Fifteenth Annual Conference of the International Speech
  Communication Association}.

\bibitem[{Rehurek and Kolkus(2009)}]{rehurek:2009language}
Radim Rehurek and Milan Kolkus. 2009.
\newblock Language identification on the web: Extending the dictionary method.
\newblock In \emph{International Conference on Intelligent Text Processing and
  Computational Linguistics}, pages 357--368. Springer.

\bibitem[{Rijhwani et~al.(2017)Rijhwani, Sequiera, Choudhury, Bali, and
  Maddila}]{rijhwani:2017estimating}
Shruti Rijhwani, Royal Sequiera, Monojit Choudhury, Kalika Bali, and
  Chandra~Shekhar Maddila. 2017.
\newblock Estimating code-switching on twitter with a novel generalized
  word-level language detection technique.
\newblock In \emph{Proceedings of the 55th Annual Meeting of the Association
  for Computational Linguistics (Volume 1: Long Papers)}, volume~1, pages
  1971--1982.

\bibitem[{Santoro et~al.(2016)Santoro, Bartunov, Botvinick, Wierstra, and
  Lillicrap}]{santoro:2016one}
Adam Santoro, Sergey Bartunov, Matthew Botvinick, Daan Wierstra, and Timothy
  Lillicrap. 2016.
\newblock One-shot learning with memory-augmented neural networks.
\newblock \emph{arXiv preprint arXiv:1605.06065}.

\bibitem[{Solorio et~al.(2014)Solorio, Blair, Maharjan, Bethard, Diab, Ghoneim,
  Hawwari, AlGhamdi, Hirschberg, Chang et~al.}]{solorio:2014overview}
Thamar Solorio, Elizabeth Blair, Suraj Maharjan, Steven Bethard, Mona Diab,
  Mahmoud Ghoneim, Abdelati Hawwari, Fahad AlGhamdi, Julia Hirschberg, Alison
  Chang, et~al. 2014.
\newblock Overview for the first shared task on language identification in
  code-switched data.
\newblock In \emph{Proceedings of the First Workshop on Computational
  Approaches to Code Switching}, pages 62--72.

\bibitem[{Sridhar and Sridhar(1980)}]{sridhar:1980syntax}
Shikaripur~N Sridhar and Kamal~K Sridhar. 1980.
\newblock The syntax and psycholinguistics of bilingual code mixing.
\newblock \emph{Canadian Journal of Psychology/Revue canadienne de
  psychologie}, 34(4):407.

\bibitem[{Sutskever et~al.(2013)Sutskever, Martens, Dahl, and
  Hinton}]{sutskever:2013importance}
Ilya Sutskever, James Martens, George Dahl, and Geoffrey Hinton. 2013.
\newblock On the importance of initialization and momentum in deep learning.
\newblock In \emph{International conference on machine learning}, pages
  1139--1147.

\bibitem[{Sutskever et~al.(2011)Sutskever, Martens, and
  Hinton}]{sutskever:2011generating}
Ilya Sutskever, James Martens, and Geoffrey~E Hinton. 2011.
\newblock Generating text with recurrent neural networks.
\newblock In \emph{Proceedings of the 28th International Conference on Machine
  Learning (ICML-11)}, pages 1017--1024.

\bibitem[{Vinyals et~al.(2016)Vinyals, Blundell, Lillicrap, Wierstra
  et~al.}]{vinyals:2016matching}
Oriol Vinyals, Charles Blundell, Tim Lillicrap, Daan Wierstra, et~al. 2016.
\newblock Matching networks for one shot learning.
\newblock In \emph{Advances in Neural Information Processing Systems}, pages
  3630--3638.

\bibitem[{Vyas et~al.(2014)Vyas, Gella, Sharma, Bali, and
  Choudhury}]{vyas:2014pos}
Yogarshi Vyas, Spandana Gella, Jatin Sharma, Kalika Bali, and Monojit
  Choudhury. 2014.
\newblock Pos tagging of english-hindi code-mixed social media content.
\newblock In \emph{Proceedings of the 2014 Conference on Empirical Methods in
  Natural Language Processing (EMNLP)}, pages 974--979.

\end{thebibliography}
\bibliographystyle{acl_natbib}

\end{document}